\documentclass{article}

\usepackage[preprint]{Styles/neurips_2025}

\workshoptitle{AI for Music}

\usepackage[utf8]{inputenc} 
\usepackage[T1]{fontenc}    
\usepackage{hyperref}       
\usepackage{url}            
\usepackage{booktabs}       
\usepackage{amsfonts}       
\usepackage{nicefrac}       
\usepackage{microtype}      
\usepackage{xcolor}         
\usepackage{amsmath}

\usepackage{graphicx}

\title{No Encore: Unlearning as \\ Opt-Out in Music Generation}

\definecolor{awcolor}{rgb}{1,0.4,0}
\definecolor{jjcolor}{rgb}{0,0.6, 0.1}
\definecolor{thcolor}{rgb}{0,0.4, 1.0}

\author{
  Jinju Kim$^{1,3}$\thanks{Work done while visiting Carnegie Mellon University.}\,
  ~~Taehan Kim$^{2,3*}$\,
  ~~Abdul Waheed$^{3}$\,
  ~~Jong Hwan Ko$^{1}$\,
  ~~Rita Singh$^{3}$ \\
  $^{1}$Sungkyunkwan University \quad
  $^{2}$Sogang University \quad
  $^{3}$Carnegie Mellon University \\
  \texttt{\normalsize \{jinjukim,abdulw,rsingh\}@andrew.cmu.edu~~taehan@sogang.ac.kr~~jhko@g.skku.edu}
}

\begin{document}

\maketitle

\begin{abstract}
    AI music generation is rapidly emerging in the creative industries, enabling intuitive music generation from textual descriptions.
    However, these systems pose risks in exploitation of copyrighted creations, raising ethical and legal concerns.
    In this paper, we present preliminary results on the first application of machine unlearning techniques from an ongoing research to prevent inadvertent usage of creative content. 
    Particularly, we explore existing methods in machine unlearning to a pre-trained Text-to-Music (TTM) baseline and analyze their efficacy in unlearning pre-trained datasets without harming model performance.
    Through our experiments, we provide insights into the challenges of applying unlearning in music generation, offering a foundational analysis for future works on the application of unlearning for music generative models. 
\end{abstract}

\section{Introduction}
With remarkable breakthroughs in generative AI, high-quality content creation has become widely available in domains with only a precise design of text input prompt~\cite{dall_e, chatgpt, valle2, sora, kim2025ossl, lee2025dittotts}. 
While advances enable music creation \cite{donahue2024hookpad}, protection of creators’ rights has become an urgent industry concern \cite{Quintais2025GenerativeAI, copyright_article12}. Problems range from unauthorized use of copyrighted material, fairness in compensation, and infringement. With European Union's AI Act and General Data Protection Regulation (GDPR), these problems are further amplified \cite{EU_GDPR_2016}. Hence, major AI music generation systems and open data platforms are beginning to establish safeguards against unauthorized use of copyrighted material for AI training \cite{harmonycloak, Sony2024, WMG2024AIStatement, WSJ2025AILicensing}. Musicians have also voiced strong opposition to the use of copyrighted works for AI without permission \cite{Reuters2025SilentAlbum, FT2025HelpIsComing}. This reflects a growing trend: while the industry acknowledges AI’s potential as a creative tool, it emphasizes that innovation must not come at the expense of artists’ ownership and copyright protection.



As music generation service provider, it is crucial to ensure that artists who opt-out of are not exploited during pre-training and after model deployment. To address these challenges, machine unlearning (MU), a method designed to selectively remove the knowledge of certain datasets from a model’s parameters can serve as a solution \cite{towards@cao}. Instead of having to retrain a large model, MU ensures that information derived from specific datasets no longer affects the model’s outputs with fine-tuning. 
MU has been explored in various domains of generative AI for its potential of removing unwanted dataset, concept, and output without harming generative performance \cite{gandikota2023erasing,thudi2022unrolling, gu2024second, liu2024machine, fan2024salun,kim2025do}.
By integrating MU techniques into music AI, a solution for ethical and compliant music generation can be offered.

This study is the first to explore machine unlearning in Text-to-Music, establishing a foundation for future research on applying unlearning to musical content generation. Our experimental results reveal the limitations of existing unlearning methods and analysis that point toward directions for safe music generation.

\section{Related Works}

\paragraph{Text-to-Music Generation}
Text-to-Music (TTM) models \cite{musicgen, li2024qamdtqualityawaremaskeddiffusion, fei2024fluxplaysmusic, chowdhury2024melfusion, agostinelli2023musiclmgeneratingmusictext, Novack2025Presto} aim to generate music that aligns with provided textual descriptions, allowing users to specify musical characteristics such as genre, mood, instrumentation, and style. 
A common practice in developing these models involves training on large-scale corpora that often contain copyrighted material, frequently sourced from platforms like YouTube or other web repositories. 
This reliance on potentially sensitive and proprietary audio data raises ethical and legal questions.




\paragraph{Generative Unlearning}


Generative unlearning applies the principles of MU to generative AI to remove the influence of particular data or concepts from the generated distribution while preserving the model’s ability to produce high-quality, diverse outputs \cite{golatkar2020eternal}. 
Approaches often involve modifying or fine-tuning a pre-trained parameters to “forget” certain patterns without significantly degrading performance on the rest of the data distribution. These techniques have been explored in vision, language, and speech domains
\cite{seo2024generative, kim2025do}.
Music, however, remains underexplored in machine unlearning. Our work represents a preliminary step toward extending generative unlearning frameworks to the complex and underexplored terrain of music generation. Please see Appendix~\ref{appx:related_works} for extended related works. 

\section{Machine Unlearning for Music Generation}

\begin{figure}
    \centering
    \includegraphics[width=1.0\linewidth]{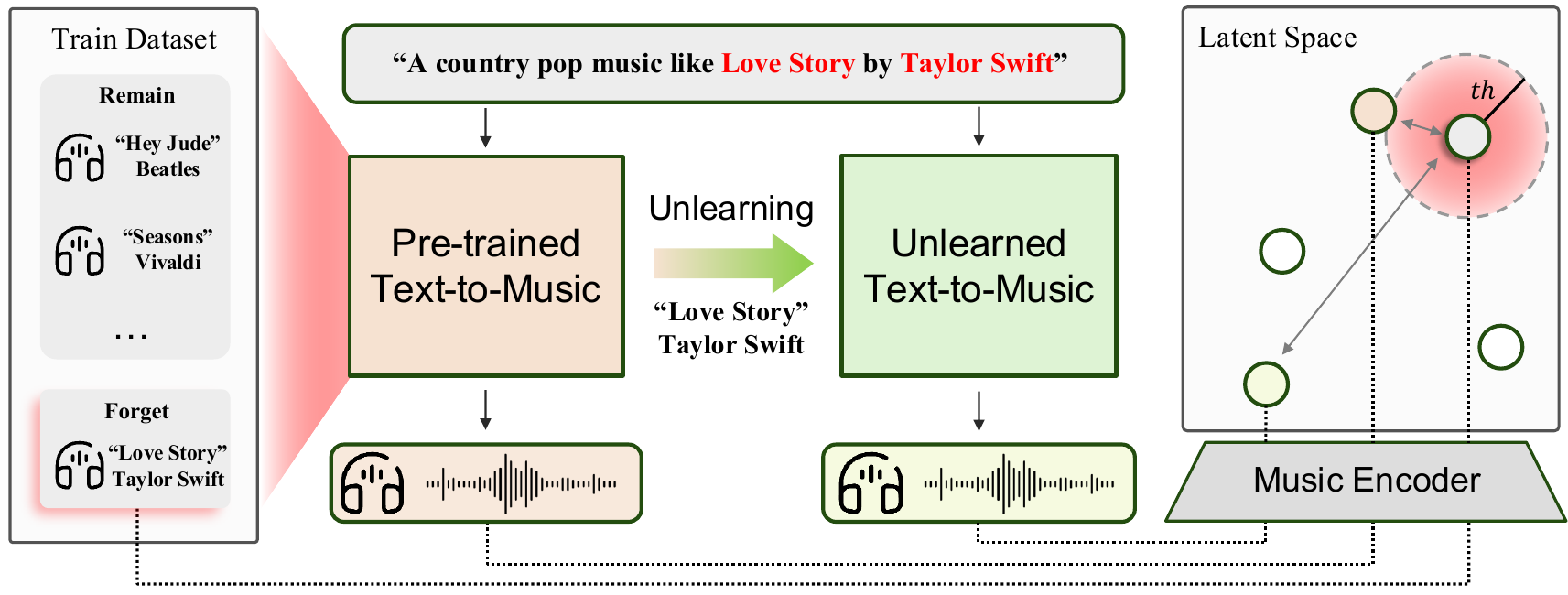}
    \caption{An overview of unlearning and its role in music generation: when a request is made to remove a specific piece of music from the pre-training dataset (the forget set), unlearning provides a potential solution. The goal is not only to erase the model’s reliance on that data, but also to ensure that future generations diverge from the copyrighted material to ensure safe music generation.}
    \label{fig:1_framework}
\end{figure}

We investigate machine unlearning in the context of music generation with the goal of removing the influence of specific training subsets while preserving model performance. Our study is guided by the following question: \textit{Can existing machine unlearning methods be directly applied to erase pre-trained musical knowledge in music generation models?} 

\subsection{Problem Formulation}  
In Figure \ref{fig:1_framework}, we consider a pre-trained TTM model parameterized by $\theta$, which maps a natural language prompt $x$ to a corresponding musical output $y$. Given a forget set $\mathcal{F} = {(x^f, y^f)}$ consisting of text-music pairs that should no longer influence the model's behavior, our goal is to modify the parameters of $\theta$ such that the model effectively "forgets" this data. Concretely, we require that:
\begin{itemize}
    \item For a forget prompt $x^f$ that targets $y^f$ in the forget set, the updated model $\theta^-$ should no longer generate the corresponding music.
    \item For all other prompts $x \notin \mathcal{F}$, the updated model $\theta^-$ should preserve the original generation performance of $\theta$. We refer to these as remain set denoted by $x^r \in \mathcal{R}$.
\end{itemize}

\subsection{Unlearning Methods}  

In this paper, we explore two representative machine unlearning approaches for TTM on MusicGen \cite{musicgen}. See Appendix~\ref{appx:exp_model} for details on the model.

Following prior work \cite{thudi2022unrolling}, we implement \textbf{Gradient Ascent (GA)}, or Negative Gradient (NG) \cite{thudi2022unrolling, feng-etal-2024-fine, gu2024second}, which maximizes the training loss for $(x^f, y^f) \in \mathcal{F}$. By pushing $\theta$ in the direction opposite to standard training, GA attempts to revert learned influence from $\mathcal{F}$. 
We also implement \textbf{Random Labeling (RL)} which aims to break the learned associations of text prompts $x^f$ and corresponding target $y^f$. RL reassigns new random targets $\tilde{y}$ to $x^f$ and thus destroys the original correspondence of $(x^f, y^f)$. This forces the model to disregard the pre-trained alignment between text prompts and their associated musical outputs \cite{golatkar2020eternal}.





\section{Experiments}

\subsection{Experimental Setups}
\paragraph{Backbone}
Experiments are performed on MusicGen on the standard hyperparameter settings from the original paper \cite{musicgen}. MusicGen is pre-trained with 20k hours of licensed music comprised of internal dataset, ShutterStock\footnote{https://www.shutterstock.com/}, and Pond5\footnote{www.pond5.com} music data. Please refer to Appenndix \ref{appx:exp} for details.


\paragraph{Datasets}
To remove pre-trained knowledge, we utilize approximately 500 samples from Pond5 dataset as the forget set. MusicCaps dataset \cite{agostinelli2023musiclmgeneratingmusictext} is employed as the remain set. Both datasets consist of extensive music-text pairs spanning a wide range of genres and styles. In our study, MusicCaps serves as $\mathcal{R}$ while Pond5 serves as $\mathcal{F}$. Specifically, we assess the preservation of performance on $\mathcal{R}$ to evaluate how unlearning with $\mathcal{F}$ affects the model's ability to generate text-coherent and diverse music. Please refer to Appendix~\ref{appx:exp_data} for details.

\paragraph{Evaluation Metrics}


We assess how each unlearning methods affect the MusicGen in three key objectives, following benchmark of MusicCaps\cite{agostinelli2023musiclmgeneratingmusictext}; plausibility, alignment, and conceptual shifts. For plausibility, we adopt Fréchet Audio Distance (\textbf{FAD}) \cite{kilgour2019frechetaudiodistancemetric} using Tensorflow's VGGish model. For alignment, we evaluate using Kullback-Leibler Divergence (\textbf{KL}) with state-of-the-art audio classifier PaSST \cite{koutini2021efficient}. Lastly, conceptual shift is evaluated to assess how coherent the generated music remains with the text prompt using Contrastive Language-Audio Pretraining score (\textbf{CLAP}) \cite{elizalde2023clap}.

\subsection{Results}

We present quantitative results to evaluate the impact of unlearning methods on retain and forget sets. Ideally, a well unlearned model should maintain its music generation performances on the remain set while selectively performing worse -- failing to generate compliant music -- when requested to generate audio with forget set. 


\paragraph{Can existing machine unlearning methods erase pre-trained musical knowledge?}

\begin{table}[h]
\centering
\caption{Quantitative results on Pond5 forget set. For the forget set, we expect degraded performance as denoted by arrows.}
\begin{tabular}{lccc}
\toprule
Method & FAD (↑) & KL (↑) & CLAP (↓) \\
\midrule
Original & 3.334 & \textbf{1.229} & 0.349 \\ \midrule
Gradient Ascent (GA) & 3.866 & 1.158 & \textbf{0.351} \\
Random Labeling (RL) & \textbf{3.875} & 1.227 & 0.320 \\
\bottomrule
\end{tabular}
\label{table:results_pond5}
\end{table}

In Table~\ref{table:results_pond5}, we evaluate models updated through unlearning on the Pond5 forget set. In terms of musical plausibility, FAD indicates that both GA and RL achieve partial forgetting: the music generated by unlearned models sounds less natural and exhibits lower audio quality. While unlearning is generally expected to reduce performance across all metrics, we observe the opposite trend for KL. This suggests that, although the unlearned models generate audio with degraded fidelity, their outputs remain similar to the forget set in terms of broad musical class (e.g., pop music). For CLAP scores, GA increases text–music alignment, whereas RL decreases it. GA appears unable to disentangle fidelity from semantic alignment. It primarily produces less natural music while still leaning on pre-trained knowledge. In contrast, RL maps prompts to essentially random outputs, teaching the model to ignore the text prompt. However, given the decrease in KL, we infer that RL only erases a limited set of anchor concepts while the model continues to retain some target knowledge.


\paragraph{Can existing machine unlearning methods preserve music generative performances?}

\begin{table}[h]
\centering
\caption{Quantitative results on MusicCaps remain set. For the remain set, an ideal unlearned model is desired to perform well. The scores are consistently lower on remain set because the model was never trained on this dataset, whose distribution differs from Pond5.}
\begin{tabular}{lccc}
\toprule
Method & FAD (↓) & KL (↓) & CLAP (↑) \\ \midrule
Original & \textbf{4.905} & \textbf{1.445} & \textbf{0.280} \\ \midrule
Gradient Ascent (GA) & 9.458 & 1.777 & 0.226 \\
Random Labeling (RL) & 9.933 & 1.696 & 0.239 \\
\bottomrule
\end{tabular}
\label{table:results_musiccaps}
\end{table}

In Table~\ref{table:results_musiccaps}, we report results on the MusicCaps remain set. An ideal unlearned model should preserve its knowledge on the remain set, maintaining performance comparable to the original model. For unlearning methods, the FAD score increases substantially, indicating that GA and RL degrade the overall quality of generated music, particularly on distributions unseen during training. Furthermore, KL increases while CLAP decreases, suggesting that unlearning significantly diminishes both the fidelity and prompt alignment of the generated outputs.

\section{Discussion}
As we pioneer unexplored questions of unlearning in music, it is critical to reflect on the limitations of current approaches, especially in the context of musical copyrights and safe generative practices. Music is inherently complex, combining melody, harmony, rhythm, timbre, and performance, all of which interact in subtle and subjective ways. Determining whether a generated piece constitutes permissible stylistic inspiration or impermissible replication is far from straightforward.

For machine unlearning to be meaningfully applied in this domain, we must be able to distinguish which features of music should be retained (e.g., general compositional ability) and which should be removed (e.g., protected melodies, identifiable stylistic traits). We argue and seek progress in unlearning for music above collaboration with musical professionals and copyright experts to define the boundaries of creative influence and infringement. Any unlearning framework must navigate these domain-specific nuances to ensure that removing unwanted content does not diminish the model’s creative potential, but instead supports a more responsible ecosystem for AI-assisted music creation.

\section{Conclusion and Future Work}
We present preliminary results of an ongoing study, the first systematic exploration of applying machine unlearning techniques to music generation. The motivation is centered on reflecting recent trends in AI copyright regulations and necessity of IP protection in music generation. We aimed to demonstrate the feasibility of removing knowledge of specific datasets from model parameters to prevent the generation of plagiarized music.

Our experiments indicate that existing machine unlearning methods are not effective when directly applied to Text-to-Music generation models. Unlearning results in model performance degradation for remain set in all evaluation measures. Additionally, our work highlights the complexity of problem formulation of unlearning in music generation. Clearer definitions of unlearning targets and precise measures to evaluate successful unlearning are required for future work. Overall, these findings establish strong baselines and underscore key directions for continued research on applying machine unlearning for music.





\small
\bibliographystyle{unsrt}
\bibliography{Styles/reference}

\appendix


\section{Experimental Details}\label{appx:exp}

\subsection{Model Configurations}\label{appx:exp_model}

\paragraph{MusicGen} MusicGen is a single-stage autoregressive Transformer‑based music generation model \cite{musicgen}. It operates directly over compressed discrete audio tokens—derived via an audio compression model—and is capable of producing high-quality music conditioned on text descriptions or melodic input. By using an efficient token interleaving strategy, MusicGen bypasses the need for multi-stage architectures or cascading models, enabling fast and controllable generation of mono or stereo audio in a single forward pass. We utilize the \texttt{MusicGen-small} variant under constrained computational resources. 

Notably, MusicGen’s training data primarily comprises audio purchased from royalty-free music platforms like Shutterstock\footnote{https://www.shutterstock.com/} and Pond5\footnote{https://www.pond5.com}. This careful selection of legally licensed content distinguishes it from many TTM approaches that rely on web-crawled, potentially copyrighted music. As a result, MusicGen offers a suitable foundation for exploring unlearning strategies that aim to remove specific training influences while maintaining lawful and ethically sourced datasets.

\paragraph{Hyperparameters} Our experiments are based on the standard hyperparameter settings from MusicGen, with specific adjustments to the learning rate and training duration to align with our unlearning objectives. We adopt a batch size of 6 without gradient accumulation, leveraging NVIDIA A100 (80GB) and RTX 4090 GPUs for training.  To accommodate the smaller batch size and ensure stable optimization, we reduce the learning rate from $1 \times 10^{-3}$ to $1 \times 10^{-4}$. 

\paragraph{Unlearning Implementations}
For the GA unlearning method, we conduct 10,000 training steps. For RL, we perform 1,000 training steps. We halt at these steps as further training drastically degrades model performance with explosive loss, referred to as catastrophical forgetting in the unlearning domain. This experimental setup enables us to effectively evaluate the impact of our unlearning strategies on both the target and unseen forget sets, ensuring that the model maintains its performance on the retained dataset despite the unlearning processes applied.

\subsection{Data Selection}\label{appx:exp_data}

\paragraph{Forget Dataset}
We utilize approximately 500 samples, representing 0.1\% of Pond5’s free trial dataset, for our unlearning experiments. The 500 samples were selected randomly using systematically designed crawling.

\paragraph{Remain Dataset}
For the remain set, we employ the MusicCaps dataset \cite{agostinelli2023musiclmgeneratingmusictext}, which comprises extensive music-text pairs spanning a wide range of genres and styles. MusicCaps was utilized by MusicGen for evaluation purposes, providing a robust benchmark for assessing generative performance and text-audio alignment. In our study, MusicCaps serves as the remain set, allowing us to measure how well the model maintains its generative capabilities and alignment with textual prompts following the unlearning process. Specifically, we assess the preservation of performance on $\mathcal{R}$ to ensure that the unlearning of $\mathcal{F}$ does not adversely affect the model’s ability to generate coherent and diverse music in response to various textual descriptions.

\section{Related Works}\label{appx:related_works}


\subsection{Generative Unlearning}

Traditional unlearning methods strive for a theoretically grounded solution, attempting to remove the influence of $\mathcal{F}$ without extensive retraining for large models. A retrained model that requires a new training from scratch is referred to as Exactly Unlearned model. Many approaches seek closed-form parameter updates or leverage stored intermediate values from the original training process, aiming to restore parameters to a state equivalent to having never seen $\mathcal{F}$. While achieving true exactness is challenging, these methods represent a direction to approximate such parameter updates. For instance, methods such as Gradient Ascent attempt to reverse the impact of $\mathcal{F}$ by directly pushing the model’s parameters away from the patterns learned from this data \cite{thudi2022unrolling}. This involves maximizing the loss on $\mathcal{F}$, thereby reducing the model’s reliance on it. Random Labeling (RL) aims to break the learned associations derived from $\mathcal{F}$ by randomly assigning new labels to those samples \cite{golatkar2020eternal}. Techniques utilizing pruning remove or modify certain parameters or subnetworks within a model that are believed to encode the information related to $\mathcal{F}$ \cite{pochinkov2024dissecting}. By systematically identifying and pruning weights associated with the forgotten data, the model’s capacity to reproduce $\mathcal{F}$ content is reduced. Although pruning may affect overall capacity, when carefully applied, it can isolate and remove targeted knowledge while preserving essential functionality.


In the vision domain, generative unlearning techniques have been explored to erase specific objects or styles from image synthesis models, ensuring that models no longer produce unwanted content while still generating plausible images \cite{seo2024generative,fan2024salun}. 
Similarly, in natural language processing, generative unlearning has been applied to large language models to remove sensitive textual patterns or biased ideas, allowing the models to remain fluent and coherent while mitigating undesirable language use \cite{gu2024second}. While research remains relatively underexplored in the audio domain, a work to remove speaker identities from zero-shot text-to-music generation models have emerged \cite{kim2025do}.



\end{document}